\documentclass{article}
\usepackage{spconf,amsmath,graphicx}

\usepackage{dsfont} % require for using \mathds{1}
\usepackage{color}

\usepackage{hyperref}
\usepackage{url}
\usepackage{dsfont} % require for using \mathds{1}
\usepackage{xspace, amsmath, amssymb, amsthm, bbm, bm}
\usepackage{color}

\usepackage{pdfsync}
\pdfoutput=1

\usepackage{graphicx}
\usepackage{tabularx}

\usepackage{algorithm}
\usepackage{algorithmic}

\usepackage{multirow}% http://ctan.org/pkg/multirow
\usepackage{hhline}% http://ctan.org/pkg/hhline

\usepackage{booktabs}% to use \midrule for drawing a horizontal line in align environment

\usepackage{capt-of}

\def\bfu{{{\mathbf u}}}
\def\barbfu{{\overline{\mathbf u}}}

\def\eps{\ensuremath{\epsilon}\xspace}
\def\expt{\ensuremath{\mathds{E}}}

\allowdisplaybreaks % This allows breaking multi-line equations

\def\one{\ensuremath{\mathds{1}\xspace}} 
 
\def\larrow{\ensuremath{\leftarrow}\xspace} 
\def\rarrow{\ensuremath{\rightarrow}\xspace} 
\def\expt{\ensuremath{\mathds{E}}}

\def\T{\ensuremath{\top}}  % transpose
\def\sig{\ensuremath{\sigma}\xspace}

\def\P{\ensuremath{\mathbb{P}}}

\def\eps{\ensuremath{\epsilon}\xspace}

\def\bigmid{\,\middle|\,\xspace}

\def\sm{{\ensuremath{\setminus}\xspace} }

\def\bfI{\ensuremath{\mathbf{I}}\xspace}

\newcommand{\what}[1]{ {\ensuremath{\widehat{#1}}} }

\def\sig{\ensuremath{\sigma}\xspace}

\def\bfI{\ensuremath{\mathbf{I}}}

\usepackage[export]{adjustbox}

\usepackage{pbox} % new line in a table cell -- http://tex.stackexchange.com/questions/2441/how-to-add-a-forced-line-break-inside-a-table-cell

\newcommand{\fr}[2]{ { \frac{#1}{#2} }}
\def\lt{\left}
\def\rt{\right}

\makeatletter
\newcommand{\vast}{\bBigg@{3}}
\newcommand{\Vast}{\bBigg@{4}}
\makeatother

\def\dsR{{{\mathds{R}}}}

\def\w{{{\mathbf w}}}

\def\y{{{\mathbf y}}}

\usepackage{pifont}% http://ctan.org/pkg/pifont
\newcommand{\cm}{\ding{51}}%

\def\p{{{\mathbf p}}}

\def\e{{{\mathbf e}}}
\def\u{{{\mathbf u}}}

\def\w{{{\mathbf w}}}

\newcommand{\red}[1]{{\color[rgb]{1,.1,.1}#1}}

\newcommand{\gr}[1]{{\color[rgb]{.8,.8,.8}#1}}

\usepackage{enumitem}

\def\Predict{{\textsc{Predict}}\xspace}
\def\Query{{\textsc{Query}}\xspace}
\def\Y{\ensuremath{\mathbf{Y}}\xspace} 
\def\h{{{\mathbf h}}}
\def\l{{{\boldsymbol \ell}}}
\def\W{\ensuremath{\mathbf{W}}}
\def\D{\ensuremath{\mathbf{D}}}
\def\L{\ensuremath{\mathbf{L}}}
\def\G{\ensuremath{\mathbf{G}}}

\title{Graph-Based Active Learning: A New Look at Expected Error Minimization}
\name{Kwang-Sung Jun and Robert Nowak}
\address{Wisconsin Institutes for Discovery, University of Wisconsin-Madison \\
  \texttt{kjun@discovery.wisc.edu, rdnowak@wisc.edu}}
\begin{document}
%\ninept
%
\maketitle
\vspace{-1em}
\begin{abstract}
  In graph-based active learning, algorithms based on expected error minimization (EEM) have been popular and yield good empirical performance.  
  The exact computation of EEM optimally balances exploration and exploitation.  
  In practice, however, EEM-based algorithms employ various approximations due to the computational hardness of exact EEM.
  This can result in a lack of either exploration or exploitation, which can negatively impact the effectiveness of active learning.
 We propose a new algorithm TSA (Two-Step Approximation) that balances between exploration and exploitation efficiently while enjoying the same computational complexity as existing approximations.
  Finally, we empirically show the value of balancing between exploration and exploitation in both toy and real-world datasets where our method outperforms several state-of-the-art methods.
\end{abstract}

\vspace{-.5em}

%%%%%%%%%%%%%%%%%%%%%%%%%%%%%%%%%%%%%%%%%%%%%%%%%%%%%%%%%%%%%%%%%%%%%%%%%%%%%%%%
\vspace{-.7em}
\section{Introduction}
\label{sec:intro}
\vspace{-.5em}
%%%%%%%%%%%%%%%%%%%%%%%%%%%%%%%%%%%%%%%%%%%%%%%%%%%%%%%%%%%%%%%%%%%%%%%%%%%%%%%%
%
This paper studies the problem of the graph-based active learning.
We are given a weighted undirected graph $G=(N,E)$ with nodes $N = \{1,\ldots,n\}$, edges $E$, and weights $w_{ij} = w_{ji} \ge 0, \forall i\le j,$ that are $0$ if there is no edge between $i$ and $j$.
Each node $i \in N$ has a label $Y_i \in \{1,-1\}$\footnote{
  Multi-class generalization is straightforward via the one-vs-the-rest reduction; see Section~\ref{sec:expr} for detail.
}.
Let $\l_1 \subseteq N$ be the initial labeled nodes.
Initially, an algorithm knows the labels of $\l_1$ only.
At each time step $t=1,2,\ldots$, an algorithm must perform
\begin{enumerate}[topsep=3pt,itemsep=-.5ex,partopsep=10ex,parsep=1ex]
\item \Predict:  Make label prediction $\what{Y}_i$ on each unlabeled nodes $i\not\in \l_{t}$.
  Let $\what{Y}_i := Y_i, \forall i \in \l_{t}$.
  An algorithm suffers error rate $\eps_{t} = \frac{1}{n} \sum_{i=1}^n \one\{ \what{Y}_i \neq Y_i \} $, which is \emph{unknown} to the algorithm.
\item \Query: Select an unlabeled node $q$ and query its label. Receive the label $Y_q$. Update $\l_{t+1} = \l_{t} \cup \{q\}$. % without noise.
\end{enumerate}
The goal is to achieve a low error rate while querying as few nodes as possible.
The problem \Predict is an instance of semi-supervised learning~\cite{zhu05semi} for which the seminal work of Zhu et al.~\cite{zhu03semi} has been successful and de facto standard, which we call label propagation (\textbf{LP}).
We thus focus on \Query.

There are many examples where the data is given by or constructed as a graph.
In document classification problems, two documents tend to be of the same topic when one cites the other or when they use the same keywords. A graph can be constructed based on such relations.
The graph can then be used to infer a given document's topic from the known topics of the other connected documents.  More generally, a graph can be constructed based on known similarities or dissimilarities between unlabeled examples in any machine learning application. For example, hand-written digits can be recognized efficiently through graph-based learning algorithms~\cite{zhu03semi}.
In all these examples, the edge weights in the graph carries important information on how strongly two nodes (examples) are related, which can be used to make label predictions.

One popular approach to \Query starts from an intuitive probabilistic model.
Consider the following probabilistic model for the random variable $\Y\in\{1,-1\}^n$:
\vspace{-.6em}
\begin{align} \label{bingrf}
\P(\Y = \y) 
&= \frac{1}{Z} \exp\lt(-\fr{\beta}{2} \sum_{i<j} w_{ij}(y_i - y_j)^2\rt),
\end{align}
where $Z$ is the normalization factor and $\beta > 0$ is a strength parameter.
The model prefers labelings $\y \in\{1,-1\}^n$ that vary smoothly across edges; i.e., larger weight $w_{ij}$ implies higher likelihood of $y_i=y_j$.
We refer to the model above as binary Markov random field (\textbf{BMRF}).
Note that BMRF would be equivalent to the Gaussian random field (GRF) if we relax the labels to belong to real values: $\Y \in \dsR^n$.
%due to its resemblance to the Gaussian random field~\cite{zhu03semi}, which we call GRF.  \textcolor{red}{Mabye better to call this a IIM, inhomogeneous Ising model (standard, i think), or a BRF (binary random field), and just call the Gaussian RF a GRF (standard, no?)}
%We emphasize that GRF assumes continuous labels $\Y \in \dsR^{n}$.

\begin{figure*}
{\small
\begin{center}\begin{tabular}{|c|cccccccccccccccccc|c|} \hline
   Node   & 1 & 2 & 3 & 4 & 5 & 6 & 7 & 8 & 9 & 10 & 11 & 12 & 13 & 14 & 15 & 16 & 17 & 18 & Error rate \\ \hline
True label& + & + & + & + & + & + & + & + & + &  - &  - &  - &  + &  + &  + &  + &  + &  + & \\ \hline
   ZLG    & \gr{\cm} & + & + & + & + & \cm & + & \cm & \red{\cm} &  \red{\cm} &  \gr{\cm} &  - & -  & -  & -  & -  & -  & - & 0.33 \\  
   SOpt   & \gr{\cm} & + & \red{\cm} & + & + & \cm & + & + & - & -  &  \gr{\cm} & -  &  \red{\cm} & +  & +  &  \cm &  + & + & 0.06 \\  
   BMRF & \gr{\cm} & + & + & + & + & \cm & + & \red{\cm} & + & -  &  \gr{\cm} & -  &  \red{\cm} & +  & +  &  \cm & +  & + & 0.00 \\  
TSA (\textbf{Ours})
          & \gr{\cm} & + & + & + & + & \cm & + & \red{\cm} & + & -  &  \gr{\cm} & -  &  \red{\cm} & +  & +  &  \cm & +  & +  & 0.00  \\ \hline 
\end{tabular}
\vspace{-.7em}
\caption{A linear chain example that contrasts different \Query Algorithms}
\label{fig:intro-ex}
\end{center}
}
\vspace{-2.5em}
\end{figure*}

If the labels $\Y$ truly follow BMRF with known $\beta$, given a set of observed labels of nodes $\l\subseteq N$, the expected error rate of a prediction strategy is well-defined; e.g., see~\eqref{risk-lookahead}.
Then, querying the node that minimizes the expected error in the next time step is a reasonable greedy strategy.
We refer to the query strategy above as \emph{expected error minimization} (\textbf{EEM}) principle.
We precisely define EEM in Section~\ref{sec:eem}.

EEM has been the main idea of many studies~\cite{zhu03combining,ji12avariance,ma13sigma}.
Define $\Y_\l := \{\Y_i\}_{i\in\l}$.
The challenge in EEM is to compute the posterior marginal of a node $i$ given labeled nodes $\l\subseteq N$:
\begin{align}\label{post-marg}
\P( Y_i \mid \Y_\l = \y_\l).
\end{align}
This is combinatorial; there is no known polynomial time algorithm for computing it, to our knowledge.
Resolving such a computational issue in EEM has been an active area of research.
Zhu et al.~\cite{zhu03combining} apply a simple approximation to~\eqref{post-marg} by posterior mean of GRF, which we call \textbf{ZLG}.
V-optimality (\textbf{VOpt})~\cite{ji12avariance} considers EEM under GRF instead of BMRF, which results in a closed-form solution.
$\Sigma$-optimality (\textbf{SOpt})~\cite{ma13sigma} takes the same approach as VOpt, but based on a different error notion called survey error.

Each EEM-based algorithm has an undesirable behavior.
Consider a linear chain of length 18 with edges between $i$ and $i+1$ for all $1 \le i \le 17$ with weight 1; see Figure~\ref{fig:intro-ex}.
Labels for node 1 and 11 are given as initial labels.
We denote labeled nodes by \cm\xspace where initial labels are in gray, the first two queries are in black, and the last two are in red.
Symbols +/- indicate the predicted labels by LP after 4 queries.
For the first query, an algorithm sees that there is at least one cut (edge connecting different labels) between node 1 and 11.
ZLG drills into this region and spends its next four queries in nailing down the cut.
Consequently, it does not query any node to the right side of node 11 and incurs large error; i.e., ZLG lacks exploration queries.
In SOpt, the first two queries does include exploration query (node 16).
Then, the next two queries include node 3 that does not reduce the error rate; node 8 would have reduced error.
SOpt selects queries by which \emph{nodes} have been labeled, ignoring what \emph{labels} they have.
In fact, this is the common characteristic of many graph-based active learning algorithms~\cite{gu12towards,gadde14active,guillory09label}.
This is why SOpt is not able to optimize exploitation queries, which results in higher error than other methods as we show in toy experiments in Section~\ref{sec:expr}.
VOpt shares the same issue, so we omit it here.
In contrast, the exact computation of EEM (row BMRF) balances between exploration and exploitation.

In this work, we propose a new algorithm \textbf{TSA} whose name comes from a two-step approximation to the posterior marginal~\eqref{post-marg}.
TSA improves upon both ZLG and SOpt without added computational complexity.  
The time complexity of TSA per query is $O(n^2)$, which is the same as ZLG and SOpt.
Unlike ZLG and SOpt, TSA balances between exploration and exploitation.
In a linear chain example in Figure~\ref{fig:intro-ex}, TSA finds the same queries as BMRF.
We present TSA in Section~\ref{sec:tsa} and empirical results in Section~\ref{sec:expr} where we observe that TSA outperforms baseline methods on several toy and real-world datasets.

%%%%%%%%%%%%%%%%%%%%%%%%%%%%%%%%%%%%%%%%%%%%%%%%%%%%%%%%%%%%%%%%%%%%%%%%%%%%%%%%
\vspace{-1.3em}
\section{Expected Error Minimization (EEM)}
\label{sec:eem}
\vspace{-.5em}
%%%%%%%%%%%%%%%%%%%%%%%%%%%%%%%%%%%%%%%%%%%%%%%%%%%%%%%%%%%%%%%%%%%%%%%%%%%%%%%%

Consider a probabilistic model over a $\Y \in \{1,-1\}^{n}$ such as~\eqref{bingrf}.
Given a set of labeled nodes $\l \subseteq N$ with label $\y_\l$, the optimal prediction is the Bayes decision rule 
\begin{align}\label{bayes-rule}
  \what{Y}_i(\Y_\l = \y_\l) := \arg\max_{y\in\{1,-1\}} \P(Y_i = y \mid \Y_\l = \y_\l) \;.
\end{align}
Note $\what{Y}_i(\Y_\l = \y_\l) = Y_i$ for $i\in\l$ trivially. 
We hereafter use $\what{Y}_i$ and omit $(\Y_\l = \y_\l)$ when it is clear from the context. 

Define the unlabeled nodes $\u := N \sm \l$.
We are interested in measuring the expected error rate of the Bayes decision rule after querying $q \in \u$.
Since we do not know $Y_q$ yet, we take expectation over $Y_q \in \{1,-1\}$ as well as $\{Y_i\}_{i \in \u\sm\{q\}}$.
We define the expected error after knowing the label $Y_q$ as follows, which we call \emph{lookahead zero-one risk} of node $q$: 
\vspace{-.5em}
\begin{align}\label{risk-lookahead}
 & R^{+q}(\Y_\l = \y_\l) :=   \notag\\
 &\qquad \expt_{\Y_q } \expt_{\Y_{\u\sm\{q\}}} \lt[\fr{1}{n} \sum_{i=1}^n \one\{ \what{Y}_i \neq Y_i \} \bigmid Y_q, \Y_\l = \y_\l \rt],
\end{align}
where $\what{Y}_i$ depends on $Y_q$ as well as $\Y_\l$.
We use $R^{+q}(\y_\l)$ as a shortcut for $R^{+q}(\Y_\l = \y_\l)$.

The expected error minimization (\textbf{EEM}) principle is to choose the query that minimizes the lookahead zero-one risk:
%%% BEG
% \footnote{
%   One can imagine considering multiple-step lookahead risk. Such a strategy would require impractically large computations.
% }:
%%% END
\vspace{-5pt}
\begin{align}\label{query-rule}
  \arg \min_{q \in N\sm\l} R^{+q}(\y_\l) \;.
\end{align}
%
% Define $\P_{\y_\l}(\cdot) := \P(\cdot | \Y_\l = \y_\l)$.
% To compute the lookahead risk, we first define the zero-one risk:
Define $\P_{\y_\l}(\cdot) := \P(\cdot | \Y_\l = \y_\l)$ and the \emph{zero-one risk}
% $ R( Y_q=y, \y_\l) 
% := \expt_{\Y_{\u\sm\{q\}}}\lt[ \sum_{i=1}^n \fr{1}{n}\one\{ \what{Y}_i \neq Y_i \} \bigmid Y_q = y, \Y_\l = \y_\l \rt] 
% = \fr{1}{n} \sum_{i=1}^n \lt( 1 - \max_{y'\in\{1,-1\}} \P_{Y_q=y, \y_\l}(Y_i = y') \rt)
% $.
\vspace{-5pt}
\begin{align} 
&R( Y_q=y, \y_\l) \notag  \\
&:= \expt_{\Y_{\u\sm\{q\}}}\lt[ \sum_{i=1}^n \fr{1}{n}\one\{ \what{Y}_i \neq Y_i \} \bigmid Y_q = y, \Y_\l = \y_\l \rt] \notag\\
&= \fr{1}{n} \sum_{i=1}^n \lt( 1 - \max_{y'\in\{1,-1\}} \P_{Y_q=y, \y_\l}(Y_i = y') \rt) \;. \label{risk}
\end{align}
Then,
\vspace{-5pt}
\begin{align}\label{query-rule-simple}
  R^{+q}(\y_\l) 
%&= \expt_{Y_q} R(Y_q, \y_\l) \\
&= \sum_{y\in\{1,-1\}} R(Y_q = y, \y_\l) \P_{\y_\l}(Y_q = y)  \;.
\end{align}
Notice that the key quantity is the posterior marginal distribution $\P_{Y_q=y, \y_\l}(Y_i = y')$ in computing~\eqref{risk} and $\P_{\y_\l}(Y_q = y)$ in~\eqref{query-rule-simple}.
%$\P_{\y_\h}(Y_i)$ for any $i \in N$ and $\h \subseteq N$.
An efficient computation of the posterior marginal would lead to an algorithm for \Predict due to~\eqref{bayes-rule}, and also to an algorithm for \Query due to~\eqref{query-rule}.

%%%%%%%%%%%%%%%%%%%%%%%%%%%%%%%%%%%%%%%%%%%%%%%%%%%%%%%%%%%%%%%%%%%%%%%%%%%%%%%%
\vspace{-5pt}
\section{Two-Step Approximation of Marginal}
\label{sec:tsa}
\vspace{-5pt}
%%%%%%%%%%%%%%%%%%%%%%%%%%%%%%%%%%%%%%%%%%%%%%%%%%%%%%%%%%%%%%%%%%%%%%%%%%%%%%%%

Consider BMRF defined in~\eqref{bingrf}.
Let $\L$ be the graph Laplacian defined by $L_{ij} := \one\{i=j\}(\sum_{k=1}^n w_{ik}) - w_{ij}$.
% Let $\W$ be the matrix of edge weights $w_{ij}, \forall i,j\in N,$ and $\D$ be a diagonal matrix with $D_{ii} := \sum_{j=1}^n w_{ij} $.
% Define the graph Laplacian $\L:= \D - \W$.
We rewrite~\eqref{bingrf} compactly:
\vspace{-.5em}
\begin{align} \label{bingrf-detail}
  \P(\Y = \y) 
&= \fr{1}{Z} \exp\lt(-\fr{\beta}{2} \y^\T \L \y\rt) \;.
\end{align}
For ease of exposition, we let $\beta = 1$; one can obtain results for $\beta\neq 1$ by replacing $\L$ with $\beta \L$.

Suppose we have observed the labels of nodes $\l$ as $\y_\l$.
We propose a \emph{two-step approximation} (\textbf{TSA}) to the posterior marginal distribution $\P_{\y_{\l}}(Y_k)$, which leads to a new \Query algorithm.
The key lies in the following log probability ratio approximation: $\log\fr{\P(Y_k = 1, \Y_\l=\y_\l)}{\P(Y_k=-1, \Y_\l=\y_\l)} \approx \log\fr{\mu(Y_k = 1, \Y_\l=\y_\l)}{\mu(Y_k = -1, \Y_\l=\y_\l)}$ for some $\mu(\cdot)$.
Define the sigmoid function $\sig(z) := (1+\exp(-z))^{-1}$.
Then, it follows that
\vspace{-.5em}
\begin{align*}
&\P(Y_k = 1 \mid \Y_\l = \y_\l)  \\
&= \fr{ \P(Y_k = 1, \Y_\l = \y_\l) }{ \P(Y_k = 1, \Y_\l = \y_\l) + \P(Y_k = -1, \Y_\l = \y_\l) } \\
&= \sig( \log \P(Y_k = 1, \Y_\l = \y_\l) - \log\P(Y_k = -1, \Y_\l = \y_\l)) \\
&\approx \sig(\log \mu(Y_k = 1, \Y_\l = \y_\l) - \log \mu(Y_k = -1, \Y_\l = \y_\l)) \;.
\end{align*}
We construct $\mu(Y_k = y_k, \Y_\l=\y_\l)$ as a two-step upperbound on $\P(Y_k = y_k, \Y_\l=\y_\l)$.
Define $\barbfu := \u \sm \{k\}$, the set of unlabeled nodes except $k$.
Let $A := \L_{k k} + \y_\l^\T \L_{\l\l} \y_{\l}$ and $g(\y_\barbfu) := -\lt( \fr{1}{2}\y_\barbfu^\T \L_{\barbfu\barbfu} \y_\barbfu + y_k\L_{k\barbfu}\y_{\barbfu} +  \y_\l^\T\L_{\l\barbfu}\y_\barbfu \rt)$.
We simplify $\log \P(Y_k=y_k,\Y_\l=\y_\l) = $
\vspace{-.5em}
\begin{align*}
-\log(Z) -\frac{1}{2} A - y_k\L_{k\l}\y_\l  + \log \Big( \sum_{\y_\barbfu} \exp( g(\y_\barbfu) ) \Big) \;.
\end{align*}
%%% BEG TODO for arxiv
%We provide details in the appendix.
%%% END
Note that the last term is the log-sum-exp function that is similar to the max operator.
This leads to our \textbf{first upperbound}:
\vspace{-.5em}
\begin{align*}
\log \lt( \sum_{\y_\barbfu} \exp( g(\y_\barbfu) ) \rt)
\le \max_{\y_\barbfu \in \{1,-1\}^{|\barbfu|}} g(\y_\barbfu)  + |\barbfu| \log 2.
\end{align*}
We now have an integer optimization problem, which is hard in general.
We relax the domain of $\y_{\barbfu}$ to real, which leads to our \textbf{second upperbound}:
\vspace{-.5em}
\begin{align*}
 \max_{\y_\barbfu \in \{1,-1\}^{|\barbfu|}} g(\y_\barbfu) \le \max_{\y_\barbfu \in \dsR^{|\barbfu|}} g(\y_\barbfu).
\end{align*}
We now have a concave quadratic maximization problem.
Find the closed form solution (see the supplementary material~\ref{sec:supp-solution}). Then, altogether,
\vspace{-.5em}
\begin{align*}
&\log \P(Y_k = y_k, \Y_\l = \y_\l) \\
&\quad\le -\log(Z) -\frac{1}{2} A - y_k\L_{k\l}\y_\l + \\
&\quad\quad \fr{1}{2}(y_k \L_{k\barbfu} + \y_\l^\T \L_{\l \barbfu}) \L_{\barbfu\barbfu}^{-1} (\L_{\barbfu k} y_k + \L_{\barbfu \l}\y_\l) + |\barbfu|\log 2 \\
&\quad =: \log \mu(Y_k = y_k, \Y_\l = \y_\l) \;,
\end{align*}
where $\L_{\barbfu\barbfu}^{-1} = (\L_{\barbfu\barbfu})^{-1}$.
Let $f_k := \log \mu(Y_k = 1, \Y_\l = \y_\l) - \log \mu(Y_k = -1, \Y_\l = \y_\l)$ be the decision value of node $k$.
We simplify $f_k$:
\vspace{-5pt}
\begin{align}\label{tsa-decision-val}
f_k 
&= -2 \L_{k\l}\y_\l + 2 \L_{k\barbfu}\L_{\barbfu\barbfu}^{-1}\L_{\barbfu\l}\y_\l 
\end{align}
for which we present a natural interpretation in our appendix.
Finally, compute $\P(Y_k = 1 \mid \Y_\l = \y_\l) \approx \sig(f_k)$ for all $k\not\in\l$ and perform EEM~\eqref{query-rule}.

\textbf{Computing Marginals Altogether}\quad
Note that we need to compute $\p(Y_k=1\mid\Y_\l=\y_\l) $ for every node $k\in\u$, and the matrix inversion in~\eqref{tsa-decision-val} is costly.
Denote by $\circ$ the Hadamard product and $[f_k]_{k\in\u}$ a vector whose $k$-th component has value $f_k$.
Note that the one-step covariance update rule says that $\begin{pmatrix} \L_{\barbfu\barbfu}^{-1} & 0 \\ 0 & 0  \end{pmatrix} = \L^{-1}_{\u\u} - \fr{(\L^{-1}_{\u\u})_{\cdot k}(\L^{-1}_{\u\u})_{k\cdot}}{(\L^{-1}_{\u\u})_{kk}} $, where we assume that node $k$ is the largest index among $\barbfu$, without loss of generality.
Using one-step covariance update rule, one can compute the marginals all at once with one matrix inversion (see the supplementary material~\ref{sec:supp-computing}):
\vspace{-5pt}
\begin{align}\label{tsa-decision-vec}
[f_k]_{k\in \u} 
&= - 2 \lt[ \fr{ 1 }{ (\L_{\u\u}^{-1})_{kk}} \rt]_k \circ (\L_{\u\u}^{-1} \L_{\u\l} \y_\l) \;.
\end{align}
%Then, evaluating EEM~\eqref{}the lookahead risk can be computed with $O(n)$ matrix inversions.
Evaluating EEM~\eqref{query-rule} involves computing~\eqref{tsa-decision-vec} $O(n)$ times.
Since the matrix inversion in~\eqref{tsa-decision-vec} can be performed in $O(n^2)$ using the one-step covariance update, the time complexity per query would be $O(n^3)$.
However, one can use the ``dongle node'' trick presented in Appendix A of~\cite{zhu03combining} to improve it to $O(n^2)$; see our Appendix for detail.

\begin{figure*}[!ht]
  \centering
  \begin{tabular}{ccc}
    \includegraphics[width=.30\textwidth]{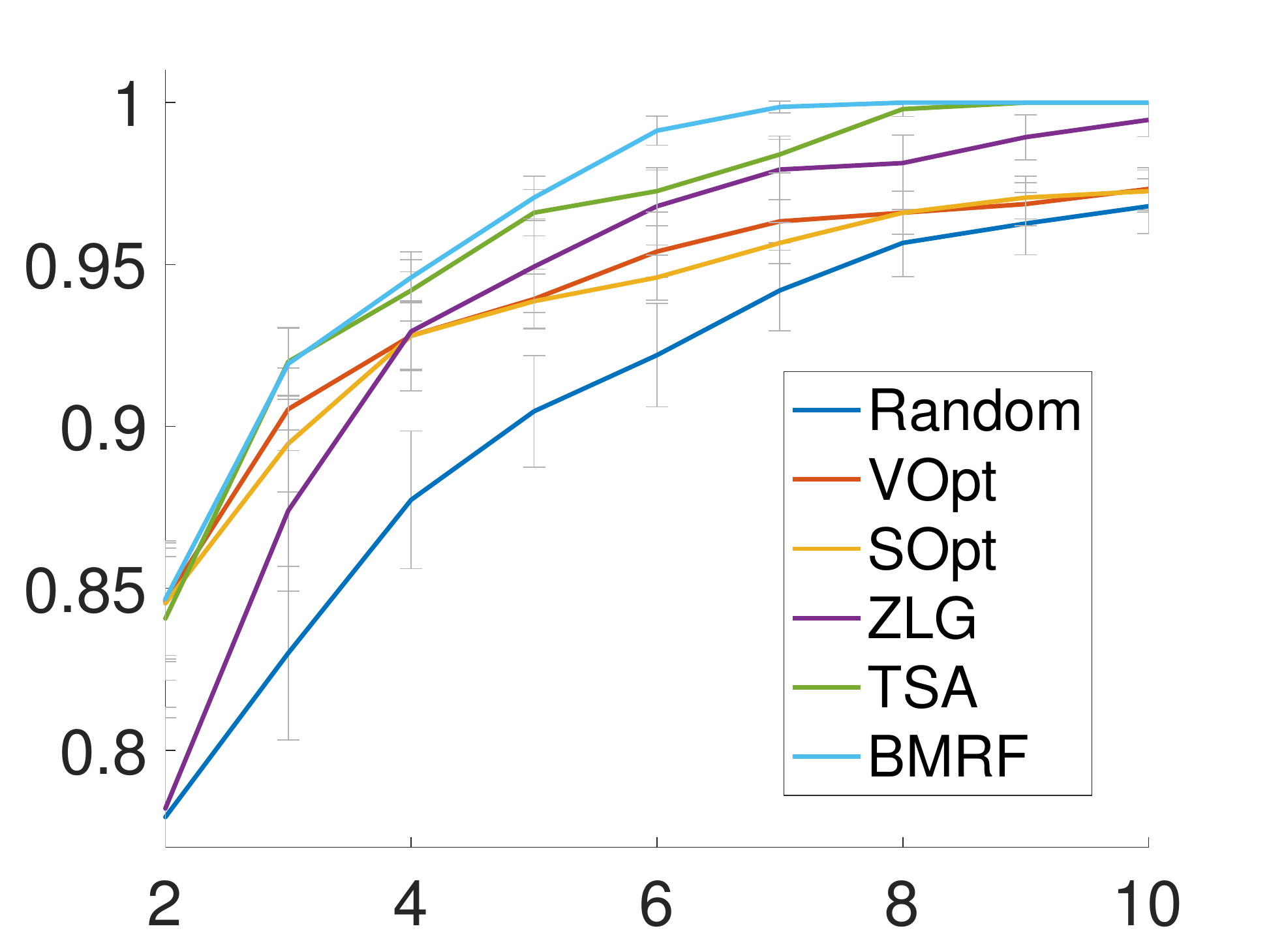} &
    \includegraphics[width=.30\textwidth]{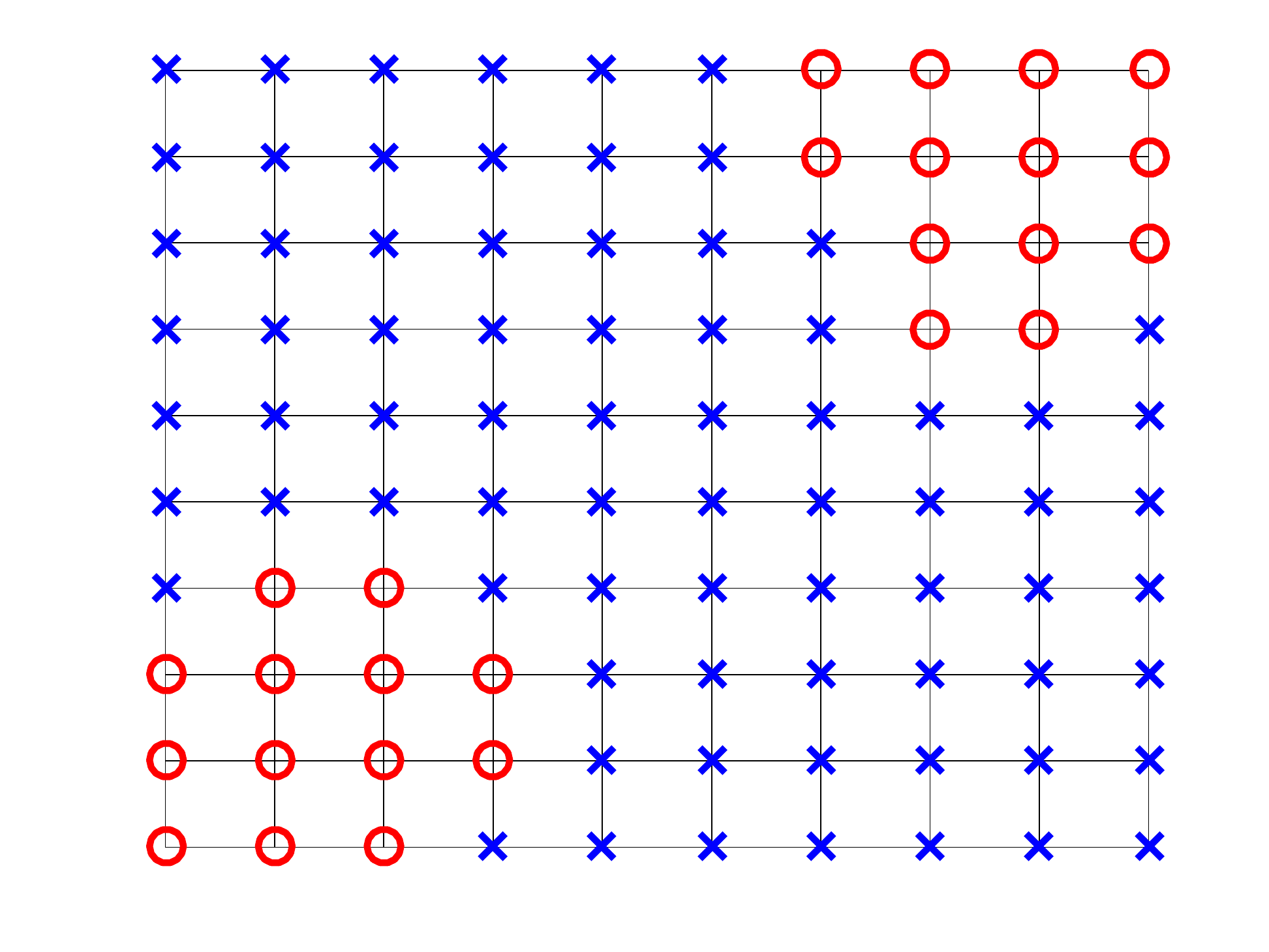} &
    \includegraphics[width=.30\textwidth]{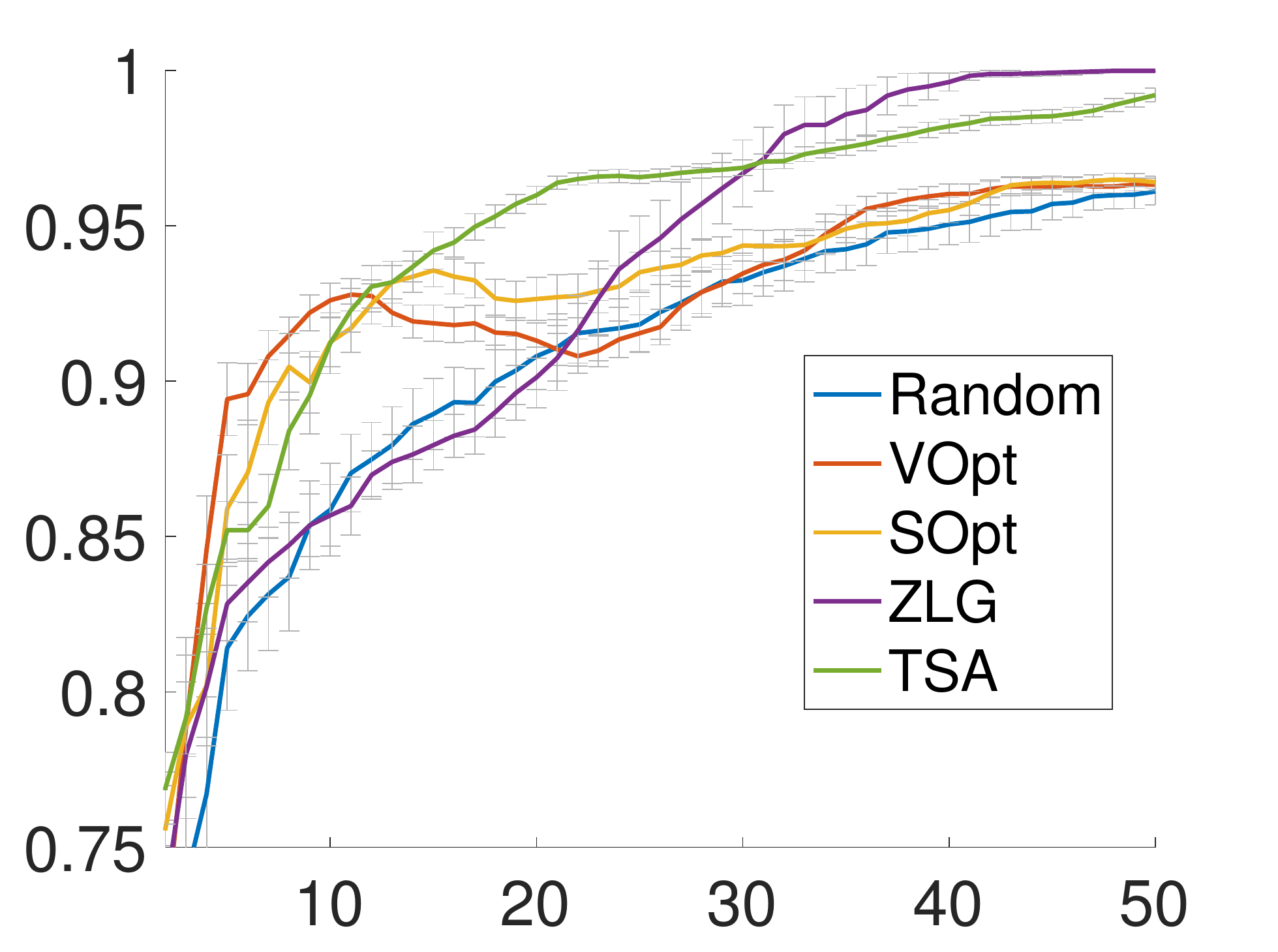} \\
    (a) Linear chain & (b) Jittered box dataset & (c) Jittered box \\
    \includegraphics[width=.30\textwidth]{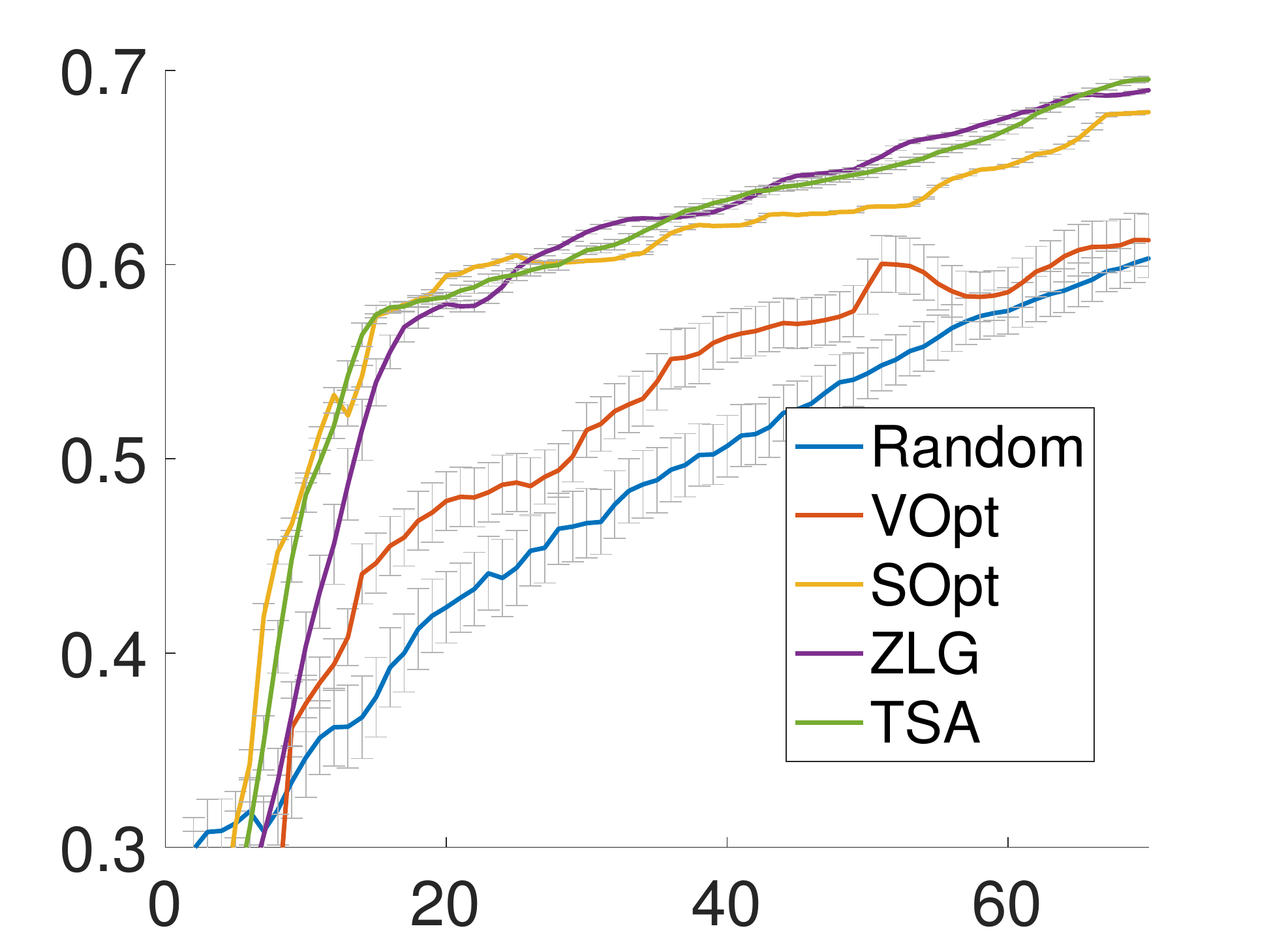} &
    \includegraphics[width=.30\textwidth]{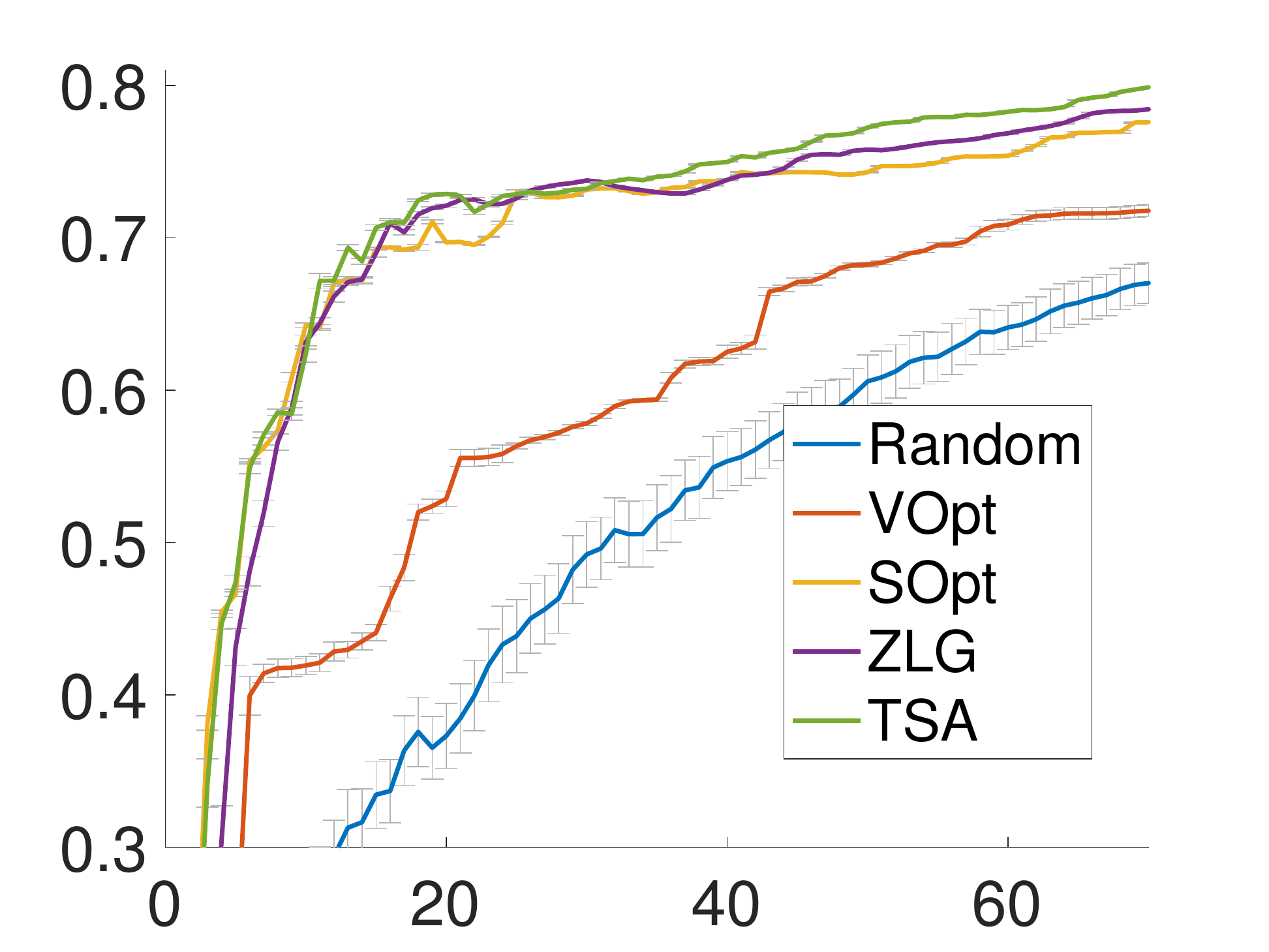} &
    \includegraphics[width=.30\textwidth]{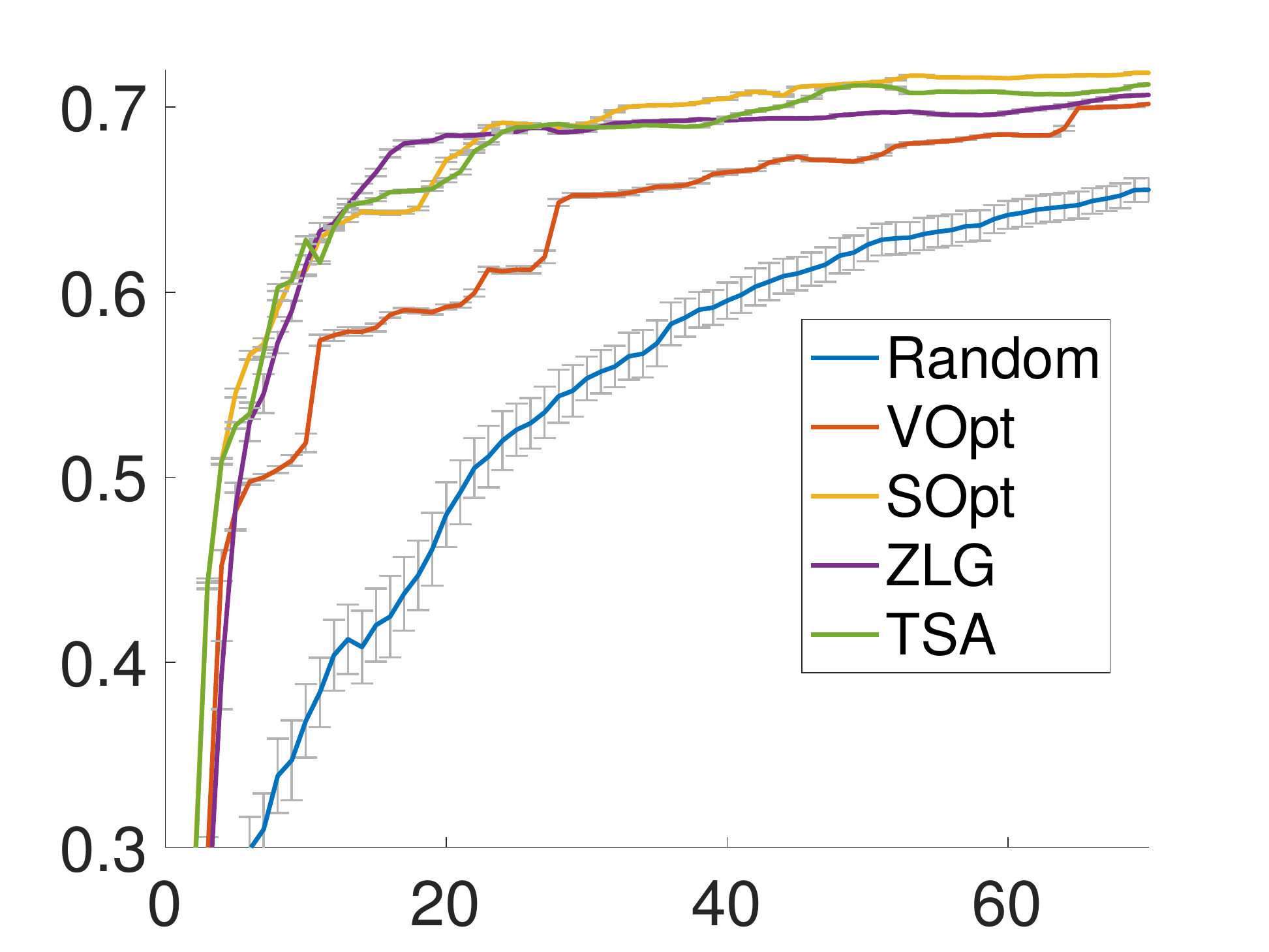} \\
    (d) DBLP & (e) CORA & (f) CITESEER \\
  \end{tabular}
  \vspace{-7pt}
  \caption{Experiment Results. Plots show accuracy vs. the number of queries. Error bars are in gray.}
  \label{fig:expr}
  \vspace{-10pt}
\end{figure*}

\textbf{Comparison to ZLG}\quad
Let $\sig^{\text{Linear}}(z) := \fr{1}{2}(z + 1)$ that is valid over $z \in [-1,1]$ only.
ZLG performs a simple approximation: $[ \P(Y_k=1\mid \Y_\l = \y_\l) ]_k \approx$
\begin{align*}
\sig^{\text{Linear}}(-\L^{-1}_{\u\u} \L_{\u\l} \y_\l) \;,
\end{align*}
where we apply $\sig^{\text{Linear}}$ elementwise.
Input to $\sig^{\text{Linear}}$ is always in $[-1,1]$ due to the property of the harmonic function~\cite{zhu03semi}.
In TSA, $[ \P(Y_k=1\mid \Y_\l = \y_\l) ]_k \approx$
%As a comparison, TSA marginal can be expressed with the logistic sigmoid function $\sig(z) = (1+\exp(-z))^{-1}$:
\begin{align*}
\sig\lt( 2 \cdot \lt[ \fr{ 1 }{ (\L_{\u\u}^{-1})_{kk}} \rt]_k \circ (-\L_{\u\u}^{-1} \L_{\u\l} \y_\l) \rt) \;.
\end{align*}
Both methods utilize $\h := (-\L_{\u\u}^{-1} \L_{\u\l} \y_\l)$, which is the decision value of LP that is thresholded at 0 to make predictions (and notice both methods lead to the same prediction).
Beside using a different sigmoid function, TSA further weights $h_k$ by $1/(\L^{-1}_{\bfu\bfu})_{kk}$ where $(\L^{-1}_{\bfu\bfu})_{kk}$ is always positive.
$(\L^{-1}_{\bfu\bfu})_{kk}$ can be interpreted as the variance of node $k$ in GRF context.
The larger the variance of a node is, the closer its decision value to 0, and the closer the marginal probability to 1/2.
Such a variance information is not utilized in ZLG.

A striking example is our introductory example in Figure~\ref{fig:intro-ex}.
When the initial labels are given for node 1 and 11, the posterior marginal $\P(Y_k=-1 | \Y_\l=\y_\l)$ for $k=12,\ldots,18$ under BMRF is (0.88, 0.79, 0.72, 0.67, 0.63, 0.60, 0.57) and under TSA is (0.88, 0.73, 0.66, 0.62, 0.60, 0.58, 0.57).
Among node 12 to 18, node 16 has the smallest lookahead risk under both methods.
However, under ZLG the marginals are (1,1,$\ldots$,1) for node 12 to 18, which results in all 0 lookahead risk.
Similarly, after querying node 6 the segment from node 2 to 5 have marginals (0,0,$\ldots$,0) and 0 lookahead risk values.
This explains why ZLG lacks exploration queries.

Another difference is that replacing occurrences of $\L$ with $\beta\L$ in ZLG results in no contribution of $\beta$ whereas in TSA there \emph{exists} contribution of $\beta$; the smaller the $\beta$ is the closer the marginal probabilities to 1/2.
We observed that $\beta$ changes the balance between exploration and exploitation.
However, parameter tuning in active learning is hard in general; we leave it as a future work and use $\beta=1$ in experiments.

%%%%%%%%%%%%%%%%%%%%%%%%%%%%%%%%%%%%%%%%%%%%%%%%%%%%%%%%%%%%%%%%%%%%%%%%%%%%%%%%
\vspace{-8pt}
\section{Experiments}
\label{sec:expr}
\vspace{-.5em}
%%%%%%%%%%%%%%%%%%%%%%%%%%%%%%%%%%%%%%%%%%%%%%%%%%%%%%%%%%%%%%%%%%%%%%%%%%%%%%%%

Throughout the experiments, all methods start from one labeled node that is chosen uniformly at random.
For every method, we break ties uniformly at random.
Let $C$ be the number of classes.
We handle multi-class case by instantiating one algorithm for each one-vs-the-rest (total $C$ runs).
After computing each one-vs-the-rest marginal (binary), we compute the multi-class marginal distribution (now multinomial) by normalizing the binary marginals.
Finally, the multi-class zero-one risk is a trivial extension of~\eqref{risk} from which we compute the EEM query.

\textbf{Toy Data} \quad
The first toy dataset is a linear chain with 15 nodes where each edge has weight 1.
We choose an edge uniformly at random and assign positive label on one side and negative on the other side.
We repeat the experiment 50 times where we assign new labels before each trial.
We plot the accuracy vs. the number of queries in Fig.~\ref{fig:expr}(a) with the confidence bounds in gray.
After 10 queries, we observe a group of methods that outperforms the rest.
This group consists of methods that are equipped with exploitation queries and thus able to nail down the exact cut.
The rest are \emph{non-adaptive} methods who are blind to observed labels.
This experiment confirms the importance of the exploitation queries.

The second toy dataset is the 10 by 10 grid graph; see Fig.~\ref{fig:expr}(b).
We assign positive labels to the 3 by 3 box at the bottom left and another one at the top right, and negative labels to the rest.
Then, for each negative nodes adjacent to a positive node, we assign positive with probability 1/2 to make the boundary ``jittered''.
We repeat the experiment 50 times where we assign new jittered labels before each trial.
We show the result in Fig.~\ref{fig:expr}(c).
There is no absolute winner.
For very early time period, both VOpt and SOpt perform slightly better than the rest since they explore only --- rough locations of the two positive boxes are discovered fast.
On the other hand, ZLG incurs very low accuracy in the first half for the following two reasons: $(i)$ before discovering a positive node, every node has the same lookahead risk and ZLG resorts to tie-breaking uniformly at random and $(ii)$ after discovering the first positive node, ZLG drills down the exact boundary of it while completely not knowing the existence of the other positive box. 
In the end, however, ZLG becomes the best since it does not waste queries on exploration.
TSA, our method, balances between exploration and exploitation and perform well on average.

\begin{table}
  \centering
\begin{tabular}{|c|c|c|c|} \hline
Name & $|N|$  & $|E|$  & The number of classes \\ \hline
DBLP &           1711 & 2898 & 4 \\
CORA &         2485 & 5069 &  7 \\
CITESEER &     2109 & 3665 & 6 \\ \hline
\end{tabular}
\vspace{-0.5em}
\caption{Real-world dataset summary}
\label{tab:dataset}
\vspace{-25pt}
\end{table}

\textbf{Real-World Data} \quad
We use exactly the same dataset as~\cite{ma13sigma}\footnote{The dataset can be download from \texttt{http://www.autonlab.org/autonweb/21763}}, which is summarized in Table~\ref{tab:dataset}.
DBLP is a coauthorship network, and both CORA and CITESEER are citation networks; see~\cite{ma13sigma} for detail.
We repeat the experiment 50 times and plot the results in Fig.~\ref{fig:expr}(d-f).
Overall, SOpt is better than ZLG for earlier time period, but ZLG is better for later time period (except in CITESEER), which we believe is due to the fact that ZLG lacks exploration queries and SOpt lacks exploitation queries, respectively. 
In contrast, TSA is as good as SOpt for earlier time period and as good as or even better than ZLG for later time period in all three datasets as TSA is able to balance between exploration and exploitation.

% For earlier time period, both SOpt and TSA are the best in all datasets while ZLG is worse than both, which is expected since ZLG lacks exploration queries.
% For later time period, SOpt is worse than both ZLG and TSA in both DBLP and CORA, which is also expected since SOpt lacks exploitation queries.
% In CITESEER, however, SOpt works well for later time period as well.

%In both DBLP and CORA, TSA is better than ZLG overall,

% \renewcommand\thesection{\Alph{section}}
% \setcounter{section}{0}
%%%%%%%%%%%%%%%%%%%%%%%%%%%%%%%%%%%%%%%%%%%%%%%%%%%%%%%%%%%%%%%%%%%%%%%%%%%%%%%%
\vspace{-.5em}
\section{Appendix}
\label{sec:appendix}
\vspace{-.5em}
%%%%%%%%%%%%%%%%%%%%%%%%%%%%%%%%%%%%%%%%%%%%%%%%%%%%%%%%%%%%%%%%%%%%%%%%%%%%%%%%

\textbf{Interpretation of the TSA Marginal}\qquad
Define $\what{\y}_\barbfu := -\L_{\barbfu\barbfu}^{-1} \L_{\barbfu\l}\y_\l $ and $\what{y}_k := - \L_{kk}^{-1} \begin{pmatrix} \L_{k\l} &\L_{k\barbfu} \end{pmatrix} \begin{pmatrix} \y_\l \\ \what{\y}_\barbfu \end{pmatrix}$.
Then, we can rewrite $f_k$:
\begin{align*}
  f_k = -2 \L_{k\l}\y_\l - 2 \L_{k\barbfu} \what{\y}_\barbfu = 2 \L_{kk} \what{y}_k \;.
\end{align*}
Recall $\P(Y_k = 1 \mid \Y_\l = \y_\l) \approx \sig(f_k)$.
The TSA marginal has the following ``imputation'' interpretation:
\begin{enumerate}[topsep=3pt,itemsep=-.5ex,partopsep=10ex,parsep=1ex] 
  \item Given labels $\y_\l$, compute the posterior mean of GRF~\cite{zhu03semi} to impute the (soft) labels of $\barbfu$: $\what{\y}_\barbfu = -\L_{\barbfu\barbfu}^{-1} \L_{\barbfu\l}\y_\l$.
  \item Based on the given labels $\y_\l$ and the imputed labels $\what{\y}_\barbfu$, compute the posterior mean of GRF of node $k$: $\what{y}_k = - \L_{kk}^{-1} \begin{pmatrix} \L_{k\l} &\L_{k\barbfu} \end{pmatrix} \begin{pmatrix} \y_\l \\ \what{\y}_\barbfu \end{pmatrix}$.
  \item Compute the TSA marginal $\P(Y_k = 1 \mid \Y_\l = \y_\l) \approx \sig(2 \L_{kk} \what{y}_k)$. 
\end{enumerate}
This reveals the close connection of TSA marginal to the posterior mean of GRF.

\textbf{Fast Computation of the Lookahead Risk}\quad
The methodology here uses the same technique presented in Appendix A of~\cite{zhu03combining}.
We summarize the result here; see the supplementary material~\ref{sec:supp-fast} for detail.
Denote by $f^{+(q,y)}_k$ the decision value of node $k$ after labeling node $q$ as $y$ so that we compute the approximation $\P(Y_k=1\mid Y_q=y, \Y_\l=\y_\l) \approx (1+\exp(-f^{+(q,y)}_k))^{-1}$.
Assume for now that $\G := \L^{-1}_{\bfu\bfu}$ and $[f_k]_{k\in\u}$ are computed from the previous iteration.
The idea is to add in a ``dongle'' node that is attached to node $q$ with weight $w_0$ with label $y$.
Then, let $\w_0 \rarrow \infty$ to arrive at
\begin{align*}
&[f^{+(q,y)}_k]_{k\in\u} =
  2 \lt[ \lt(\G_{kk} - \fr{\G_{kq}^2}{\G_{qq}}\rt)^{-1} \rt]_k \circ \\
&\qquad\lt( \fr{1}{2}\mbox{diag}(\G) \circ [f_k]_k + \lt( \fr{ y }{ \G_{qq} } - \fr{f_q}{2}\rt) \G_{\cdot q}  \rt) \;.
\end{align*}
One can verify that $f^{+(q,y)}_q = \infty$ for $y=1$ and $-\infty$ for $y=-1$, correctly. %, and thus safely remove the dimension $q$ from it.
Once we find the solution $q'$ of EEM~\eqref{query-rule} for the current time step, then we can prepare $\L^{-1}_{(\bfu\sm\{q'\})(\bfu\sm\{q'\})}$ for the next iteration using the one-step covariance update.
This implies that the time complexity per query is $O(n^2)$ for performing EEM.
Note that the full matrix inversion with cost $O(n^3)$ has to be performed for the first query.

%%%%%%%%%%%%%%%%%%%%%%%%%%%%%%%%%%%%%%%%%%%%%%%%%%%%%%%%%%%%%%%%%%%%%%%%%%%%%%%%
%%%%%%%%%%%%%%%%%%%%%%%%%%%%%%%%%%%%%%%%%%%%%%%%%%%%%%%%%%%%%%%%%%%%%%%%%%%%%%%%

%%%%%%%%%%%%%%%%%%%%%%%%%%%%%%%%%%%%%%%%%%%%%%%%%%%%%%%%%%%%%%%%%%%%%%%%%%%%%%%%
\section*{Acknowledgements}
%%%%%%%%%%%%%%%%%%%%%%%%%%%%%%%%%%%%%%%%%%%%%%%%%%%%%%%%%%%%%%%%%%%%%%%%%%%%%%%%
This work was partially supported by the National Science Foundation grants CCF-1218189
and IIS-1447449 and by MURI grant ARMY W911NF-15-1-0479.

%(\kwang{placeholder~\cite{freund97using}}).

\bibliographystyle{IEEEbib}
\bibliography{library-shared}

%%%%%%%%%%%%%%%%%%%%%%%%%%%%%%%%%%%%%%%%%%%%%%%%%%%%%%%%%%%%%%%%%%%%%%%%%%%%%%%%
\onecolumn
\begin{center}
{\Large\bf Supplementary Material}
\end{center}
\renewcommand\thesection{\Alph{section}}
\setcounter{section}{0}

\section{Derivations}

%%%%%%%%%%%%%%%%%%%%%%%%%%%%%%%%%%%%%%%%%%%%%%%%%%%%%%%%%%%%%%%%%%%%%%%%%%%%%%%%
\subsection{The Solution of the Second Upper Bound}
\label{sec:supp-solution}
%%%%%%%%%%%%%%%%%%%%%%%%%%%%%%%%%%%%%%%%%%%%%%%%%%%%%%%%%%%%%%%%%%%%%%%%%%%%%%%%
Recall that our second upperbound involves the following optimization problem:
\begin{align*}
 \max_{\y_\barbfu \in \dsR^{|\barbfu|}} g(\bfu) = 
 \max_{\y_\barbfu \in \dsR^{|\barbfu|}} -\lt( \fr{1}{2}\y_\barbfu^\T \L_{\barbfu\barbfu} \y_\barbfu + (y_k\L_{k\barbfu} + \y_\l^\T\L_{\l\barbfu})\y_\barbfu \rt) \;.
 \end{align*}
By equating its derivative to zero, $-\L_{\barbfu\barbfu} \y_\barbfu - y_k\L_{k\barbfu} - \y_\l^\T \L_{\l\barbfu} = 0$.
This leads to the solution $\y^*_\barbfu := -\L_{\barbfu\barbfu}^{-1} (\L_{\barbfu k} y_k - \L_{\barbfu\l} \y_\l) = -\L_{\barbfu\barbfu}^{-1} \begin{pmatrix} \L_{\barbfu k} & \L_{\barbfu\l}\end{pmatrix}  \begin{pmatrix} y_k \\ \y_\l \end{pmatrix} $ .
Note the solution here is equivalent to the posterior mean of the GRF given labels $\y_\l$ and $y_k$.
By plugging in $\y^*_\barbfu$ into the objective, we have
\begin{align*}
&\max_{\y_\barbfu \in \dsR^{|\barbfu|}} -\lt( \fr{1}{2}\y_\barbfu^\T \L_{\barbfu\barbfu} \y_\barbfu + (y_k\L_{k\barbfu} + \y_\l^\T\L_{\l\barbfu})\y_\barbfu \rt) \\
&\quad= \max_{\y_\barbfu \in \dsR^{|\barbfu|}} -\lt( \fr{1}{2}\y_\barbfu^\T \L_{\barbfu\barbfu} + y_k\L_{k\barbfu} + \y_\l^\T\L_{\l\barbfu}\rt)\y_\barbfu \\
&\quad= \lt(-\fr{1}{2}(y_k \L_{k\barbfu} + \y_\l^\T \L_{\l \barbfu}) + y_k \L_{k\barbfu} + \y_\l^\T \L_{\l \barbfu} \rt)\L_{\barbfu\barbfu}^{-1} (\L_{\barbfu k} y_k + \L_{\barbfu \l}\y_\l) \\
&\quad= \fr{1}{2}(y_k \L_{k\barbfu} + \y_\l^\T \L_{\l \barbfu}) \L_{\barbfu\barbfu}^{-1} (\L_{\barbfu k} y_k + \L_{\barbfu \l}\y_\l)\quad.
 \end{align*}

%%%%%%%%%%%%%%%%%%%%%%%%%%%%%%%%%%%%%%%%%%%%%%%%%%%%%%%%%%%%%%%%%%%%%%%%%%%%%%%%
\subsection{Computing Marginals Altogether}
\label{sec:supp-computing}
%%%%%%%%%%%%%%%%%%%%%%%%%%%%%%%%%%%%%%%%%%%%%%%%%%%%%%%%%%%%%%%%%%%%%%%%%%%%%%%%
Note that one needs to compute $\P(Y_k = 1\mid \Y_\l = \y_\l) $ for every $k \not\in \l$, which can be expensive.
Recall that the set of the whole unlabeled nodes is $\u = \barbfu \cup \{ k\}$.
Let $\G := \L^{-1}_{\u\u}$.
By one-step covariance update rule, if $k$ is the largest index among $\u$
$$
\begin{pmatrix} \L_{\barbfu\barbfu}^{-1} & 0 \\ 0 & 0 \end{pmatrix} = \G - \fr{\G_{\cdot k} \G_{k \cdot}}{\G_{kk}} \;.
$$
If $k$ is not the largest, the LHS has zeros for all elements of the column and row corresponding to node $k$, and removing these zeros results in $\L_{\barbfu\barbfu}^{-1}$.  
Then, 
\begin{align}
[f_k]_{k\in \u} 
&= \lt[ -2\L_{k\l}\y_\l + 2\L_{k\barbfu} \L_{\barbfu\barbfu}^{-1} \L_{\barbfu\l}\y_\l \rt]_k \notag \\
&\stackrel{(a)}{=} \lt[ -2\L_{k\l}\y_\l + 2\L_{k\u} \lt( \G - \fr{\G_{\cdot k} \G_{k \cdot}}{\G_{kk}}\rt)\L_{\u\l}\y_\l \rt]_k \notag \\
&= (-2 \bfI + 2\L_{\u\u}\G) \L_{\u\l}\y_\l - 2  \lt[\L_{k\u} \fr{\G_{\cdot k} \G_{k \cdot}}{\G_{kk}} \L_{\u\l}\y_\l \rt]_k \notag \\ 
&\stackrel{(b)}{=} - 2  \lt[\lt(\L_{k\u} \fr{ \G_{\cdot k} }{ \G_{kk} }\rt) \cdot \lt( \G_{k \cdot} \L_{\u\l}\y_\l \rt) \rt]_k \notag \\ 
&\stackrel{(c)}{=} - 2  \lt[\L_{k\u} \fr{ \G_{\cdot k} }{ \G_{kk} } \rt]_k  \circ \lt[ \G_{k \cdot} \L_{\u\l}\y_\l \rt]_k \notag \\ 
&= - 2  \lt[ \fr{ 1 }{ \G_{kk} } \rt]_k  \circ \lt( \G \L_{\u\l} \y_\l \rt) \notag \\ 
&= - 2 \cdot \lt[ \fr{ 1 }{ (\L_{\u\u}^{-1})_{kk}}\rt]_k \circ  \lt(\L_{\u\u}^{-1} \L_{\u\l} \y_\l\rt), \label{eq:decision-rule}
\end{align}
where (a) is due to the one-step covariance update rule (note the change from $\barbfu$ to $\u$), (b) is due to $\L_{\u\u} \G = \bfI$, and (c) holds by observing that it is a product of two scalars.

%%%%%%%%%%%%%%%%%%%%%%%%%%%%%%%%%%%%%%%%%%%%%%%%%%%%%%%%%%%%%%%%%%%%%%%%%%%%%%%%
\subsection{Fast computation of the lookahead risk}
\label{sec:supp-fast}
%%%%%%%%%%%%%%%%%%%%%%%%%%%%%%%%%%%%%%%%%%%%%%%%%%%%%%%%%%%%%%%%%%%%%%%%%%%%%%%%
We need to compute the lookahead risk for every unlabeled node $i \in \u = N \sm \l$.
A naive approach requires us to explicitly compute one-step lookahead inversion for each $i \in \u$, which takes $O(n^3)$ time for choosing a query.
Let $i \in \u$ be the node for which we like to compute the lookahead risk and define $\barbfu := \u \sm \{i\}$.
Let $[f_k]_{k \in \u}$ be the decision value vector after observing $\Y_\l = \y_\l$.
Let $[f^{+i}_k]_{k \in \barbfu}$ be the decision vector after observing $\Y_{\l \cup \{i\}} = \y_{\l \cup \{i\}}$.
Given $[f_k]_k$, we wish to compute $[f^{+i}_k]_k$ without explicitly computing the one-step lookahead inversion $\L_{\barbfu\barbfu}^{-1} $. 

Suppose we like to set $Y_i$ to be $y_i \in \{+1, -1\}$.
The solution starts from adding a node named $0$ with label $y_0 \larrow y_i$ to the graph and add an edge between node $0$ and $i$ with weight $w_0$ while leaving $Y_i$ unobserved.
The new node is a ``dongle'' attached to node $i$.
Let $\L^{+} := \D^{+} - \W^{+}$ be the graph Laplacian of the augmented graph.
Denote by $\e_i$ the indicator vector with $i$-th component being 1.
Then, we express the decision vector $[f^{+0}_k]_{k \in \u}$ from the augmented graph as follows (then we will later take $w_0$ to infinity to get $[f^{+i}_k]_k$):
\begin{align*}
  [f^{+0}_k]_k &= 2 \lt[ \fr{1}{((\L^+_{\u\u})^{-1})_{kk}} \rt]_k \circ \lt( (\L^{+}_{\u\u})^{-1} \W^+_{\u\l} \y_{\l \cup \{0\} } \rt) \\
   &= 2 \lt[ \fr{1}{((\L^+_{\u\u})^{-1})_{kk}} \rt]_k \circ \lt( (w_0 \e_i \e_i^\T + \D_{\u\u} - \W_{\u\u})^{-1} (w_0 y_0 \e_i + \W_{\u\l} \y_{\l}) \rt) \\
   &= 2 \lt[ \fr{1}{((\L^+_{\u\u})^{-1})_{kk}} \rt]_k \circ \lt( (w_0 \e_i \e_i^\T + \L_{\u\u})^{-1} (w_0 y_0 \e_i + \W_{\u\l} \y_{\l}) \rt) \;.
\end{align*}
Let $\G := \L_{\u\u}^{-1}$.
Applying the matrix inversion lemma, $(w_0 \e_i \e_i^\T + \L_{\u\u})^{-1} = \G - \fr{\G_{\cdot i} \G_{i\cdot}}{w_0^{-1} + \G_{ii}}$.
Then,
\begin{align*}
 \lt[ \fr{1}{((\L^+_{\u\u})^{-1})_{kk}} \rt]_k
&= \lt[ \fr{1}{ \G_{kk} - \fr{\G_{ki} \G_{ik}}{w_0^{-1} + \G_{ii}}  } \rt]_k  \\
&\stackrel{w_0 \rarrow \infty}{\longrightarrow} \lt[ \fr{1}{ \G_{kk} - \fr{\G_{ki}^2}{\G_{ii}}  } \rt]_k
\end{align*}
and
\begin{align*}
& (w_0 \e_i \e_i^\T + \L_{\u\u})^{-1} (w_0 y_0 \e_i + \W_{\u\l} \y_{\l}) \\
&= \lt( \G - \fr{\G_{\cdot i} \G_{i\cdot}}{w_0^{-1} + \G_{ii}}  \rt)  (w_0 y_0 \e_i + \W_{\u\l} \y_{\l}) \\ 
&= w_0 y_0 \G_{\cdot i} + \G \W_{\u\l} \y_\l - \fr{w_0 y_0 \G_{\cdot i}\G_{ii}}{ w_0^{-1} + \G_{ii} } - \fr{\G_{\cdot i}\G_{i\cdot} \W_{\u\l} \y_\l}{ w_0^{-1} + \G_{ii} } \\
&= \fr{ y_0 \G_{\cdot i} + w_0 y_0 \G_{\cdot i}\G_{ii} }{ w_0^{-1} + \G_{ii} } + \G \W_{\u\l} \y_\l - \fr{w_0 y_0 \G_{\cdot i}\G_{ii}}{ w_0^{-1} + \G_{ii} } - \fr{\G_{\cdot i}\G_{i\cdot} \W_{\u\l} \y_\l}{ w_0^{-1} + \G_{ii} } \\
&= \fr{ y_0 \G_{\cdot i} }{ w_0^{-1} + \G_{ii} } + \G \W_{\u\l} \y_\l - \fr{\G_{\cdot i}\G_{i\cdot} \W_{\u\l} \y_\l}{ w_0^{-1} + \G_{ii} } \\
&\stackrel{w_0 \rarrow \infty}{\longrightarrow} \fr{ y_0 \G_{\cdot i} }{ \G_{ii} } + \G \W_{\u\l} \y_\l - \fr{\G_{\cdot i}\G_{i\cdot} \W_{\u\l} \y_\l}{\G_{ii} } \\
&\stackrel{\eqref{eq:decision-rule}}{=} \fr{ y_0 \G_{\cdot i} }{ \G_{ii} } + \fr{1}{2}\mbox{diag}(\G) \circ [f_k]_k - \fr{1}{2}\G_{\cdot i} f_i \;.
\end{align*}
Therefore,
\begin{align*}
[f^{+0}_k]_k 
&\stackrel{w_0 \rarrow \infty}{\longrightarrow} 
  2 \lt[ \lt(\G_{kk} - \fr{\G_{ki}^2}{\G_{ii}}\rt)^{-1} \rt]_k \circ \lt( \fr{1}{2}\mbox{diag}(\G) \circ [f_k]_k + \lt( \fr{ y_0 }{ \G_{ii} } - \fr{f_i}{2}\rt) \G_{\cdot i}  \rt) 
= [f^{+i}_k]_{k \in \u }  \;,
\end{align*}
where one can show that $f_i^{+i} = +\infty$ if $ y_0 = 1$ and $-\infty$ if $y_0 = -1$.

\end{document}